\documentclass[10pt,twocolumn,letterpaper]{article}

\usepackage{cvpr}
\usepackage{times}
\usepackage{epsfig}
\usepackage{graphicx}
\usepackage{amsmath}
\usepackage{amssymb}

\usepackage{array}
\usepackage{color}
\usepackage{algorithm}
\usepackage{algpseudocode}

\newif\ifdraft
\drafttrue

\ifdraft
 
 \newcommand{\PF}[1]{{\color{red}{\bf PF: #1}}}
 \newcommand{\PMN}[1]{{\color{green}{\bf #1}}}
 
 \newcommand{\MS}[1]{{\color{magenta}{\bf MS: #1}}}
 
\else
 
 \newcommand{\PF}[1]{}
 
 \newcommand{\PMN}[1]{}
 
 \newcommand{\MS}[1]{}
\fi

\newcommand{\comment}[1]{}

\newcommand{\dataset}{\mathcal{D}}
\newcommand{\cdataset}{\mathcal{C}}
\newcommand{\lagrangian}{\mathcal{L}}
\newcommand{\x}{\mathbf{x}}
\newcommand{\y}{\mathbf{y}}
\renewcommand{\v}{\mathbf{v}}
\newcommand{\w}{\mathbf{w}}
\newcommand{\dw}{\mathrm{d}\mathbf{w}}
\newcommand{\C}{\mathbf{C}}
\newcommand{\B}{\mathbf{B}}

\renewcommand{\c}{\mathbf{c}}
\renewcommand{\b}{\mathbf{b}}
\newcommand{\pdv}[2]{\frac{\partial #1}{\partial #2}}
\DeclareMathOperator{\st}{s.t.}
\DeclareMathOperator*{\diag}{diag}
\DeclareMathOperator*{\argmax}{arg\,max}
\DeclareMathOperator*{\median}{median}
\DeclareMathOperator*{\Rop}{Rop}
\DeclareMathOperator*{\Lop}{Lop}
\DeclareMathOperator*{\joint}{\emph{joint}}

\cvprfinalcopy 

\def\cvprPaperID{****} 

\ifcvprfinal\fi

\begin{document}

\title{Imposing Hard Constraints on Deep Networks: Promises and Limitations}

\author{
\begin{tabular}{ccc}
Pablo M\'{a}rquez-Neila & Mathieu Salzmann & Pascal Fua\\
{\tt\small pablo.marquezneila@epfl.ch} & {\tt\small mathieu.salzmann@epfl.ch} & {\tt\small pascal.fua@epfl.ch}
\end{tabular}\\
\'{E}cole Polytechnique F\'{e}d\'{e}rale de Lausanne (EPFL)\\
}

\maketitle

\begin{abstract}

Imposing constraints on the output of a Deep Neural Net is one way to improve the quality of its predictions while loosening the requirements for labeled training data. Such constraints are usually imposed as soft constraints by adding new terms to the loss function that is minimized during training. An alternative is to impose them as hard constraints, which has a number of theoretical benefits but has not been explored so far due to the perceived intractability of the problem.

In this paper, we show that imposing hard constraints can in fact be done in a computationally feasible way and delivers reasonable results. However, the theoretical benefits do not materialize and the resulting technique is no better than existing ones relying on soft constraints. We analyze the reasons for this and hope to spur other researchers into proposing better solutions.  
\end{abstract}


\section{Introduction}

Deep Neural Networks  (DNNs) have  become fundamental  Computer Vision tools but can only give their full measure when enough training data is available. When there is not enough, good performance can still be obtained by applying well-established approaches, such as performing data  augmentation,  generating synthetic data, or  imposing constraints on the network's output. Such constraints often are regularization constraints that act as a form of weak supervision. Sometimes, they are also used to make domain-specific knowledge explicitly available to the training algorithm to leverage unlabeled data. For example, symmetry constraints are introduced in the recent CNN approach to human pose estimation of~\cite{Zhou17a}  to regularize the prediction.  Such constraints improve the quality of the DNN's predictions while loosening the requirements for labeled training data.

In the Deep Learning context, constraints are usually treated as \emph{soft} ones and imposed by adding extra penalty terms to the standard loss function. While conceptually simple,
this approach has two major drawbacks. First, it makes it necessary to
wisely choose the relative importance of the different terms in the loss function, which is not easy. Second, this gives no guarantee that the constraints will be satisfied in practice. It has been shown that, in many cases, replacing the soft constraints by hard ones addresses both of these problems~\cite{Fletcher87,Gill81}: The optimizer automatically chooses the relative weights and the constraints end up being satisfied to machine precision if a feasible solution exists. As a result, the algorithms relying on hard constraints are typically more robust and simpler to deploy. In this study, we investigate whether the same holds in the Deep Learning context. 

There is a pervasive belief in the community that imposing hard constraints on Deep Nets is impractical~\cite{pathak2015} because typical networks involve  millions of free parameters  that  need  to  be   learned,  which  tends  to  overwhelm  standard constrained optimization  algorithms. Our contribution is to show that this is not the real problem and that, by using a Krylov subspace approach, it is in fact possible to solve the very large  linear systems that these algorithms generate. As a result, our method, while slow, returns meaningful results.  

However, somewhat surprisingly, we observed that imposing soft constraints instead of hard ones yields even better results, while being far less computationally demanding.  Having studied the reasons for this counter-intuitive behavior, we have concluded that it arises from the fact that it is hard to guarantee that the linearized constraints used during the optimization are independent.  This, in turn, opens perspectives on how to overcome the problem and eventually enable us to take full advantage of the power of hard constraints in the framework of Deep Learning. 


\section{Related Work}
\label{sec:related}

Given a labeled training set $\dataset=\{(\x_i,\y_i), 1 \leq i \leq N\}$ of $N$ pairs of input vectors $\x_i$ and output vectors  $\y_i$, Deep Learning can be understood as finding a mapping $\phi(\cdot,\w)$ such that  $\phi(\x_i,\w) \approx \y_i$ for all $i$, with  $\mathbf{w}$ an $N_P$~dimensional vector of parameters to be learned. This is usually achieved  using  Stochastic Gradient  Descent (SGD)  to find
\begin{eqnarray*}
  \w^*   & = & \mbox{arg min}_{\w} R(\w) \\
  \mbox{with } R(\w) & = & \frac{1}{N}\sum_iL(\phi(\x_i, \w), \y_i) \; \; .
  \label{eq:LossU}
 \end{eqnarray*}
$R$ approximates the  expected value of the loss $L$, and is also known as the {\it risk} function.

A standard way to bias the result of this minimization is to add penalty terms to the risk function, for example to regularize the computation or introduce prior knowledge. Constraints imposed in this way are known as {\it soft} constraints because the computation yields a parameter vector $\w^*$ that achieves a compromise between minimizing the risk and satisfying the constraints. An alternative to using soft constraints is to use {\it hard} ones, which in our context means solving 
\begin{equation}
     \w^*  = \min_{\w} R(\w)  \; , \; \st \; C_{k}(\w) = 0 \mbox{ for } 1 \le k \le N_c \; , \label{eq:LossC}
\end{equation}
where the~$C_k(\cdot)$ are the $N_c$~constraints we want to enforce. 

In traditional optimization, it is well known that minimizing an objective function that  includes  many  penalty terms yields  poor   convergence   properties~\cite{Fletcher87,Gill81}: First, the optimizer is   likely to minimize the  constraint terms while ignoring the remaining terms of the objective function. Second, if one  tries to enforce  several constraints of different natures, the penalty  terms  are unlikely to be  commensurate  and one has to face the difficult problem of adequately weighing the various constraints. 

Constrained  optimization techniques offer a solution to both problems, essentially by also optimizing for the weights of the constraint terms as the optimization proceeds. In our field, this has been used to train a kernelized latent variable model to impose equality and inequality constraints~\cite{Varol12a} and to show that a Gaussian Process can be made to satisfy linear and quadratic constraints~\cite{Salzmann10c}. Similar techniques have also been proposed to constrain the output of Neural Nets before they became deep~\cite{Platt1988,Zhang1992}. 
However, the constrained optimization techniques used in these papers would not be applicable to DNNs. This is because they involve solving  large linear systems of  equations, even for moderate size problems.  When dealing with the millions of parameters that define typical Deep Learning models, the numerical techniques at the heart of these approaches would have to handle matrices whose dimensions would also be that big. This would indeed ``make a direct optimization of the constraints difficult",  as stated in~\cite{pathak2015}, if not downright impossible on available hardware.  

In fact, the approach of~\cite{pathak2015} is the only one we know of that tries to address this dimensionality problem when imposing hard constraints on a Deep Network. In this image segmentation work, the constraints are linear and used to synthesize training labelings from the network output, which avoids having to explicitly impose the constraints while minimizing the loss function. This an effective but very specialized solution. By contrast, the problem we address in this paper is that of enforcing completely generic non-linear constraints. 




\section{Formalization}
\label{sec:formal}

As discussed in Section~\ref{sec:related}, given a labeled training dataset $\dataset = \{\x_i, \y_i\}_{i=1}^{N}$ and a Deep Network architecture $\phi$ with parameters $\w$,
training a Deep Net amounts to finding the parameter $\w^*$ that minimizes the risk function $R(\w)$ of Eq.~\ref{eq:LossU}. 
In other words, $\w^*$ should ideally be such that $\forall i \leq N, \phi(\x_i,\w^*) \approx y_i$.

A specificity of Deep Learning is that it rarely, if ever, makes sense to impose hard constraints on the $\w$ parameters themselves, except for normalization that can usually be achieved more simply. Here, we therefore focus on the problem of imposing constraints on the network outputs. For example, in the experiment section we will consider 3D human body pose estimation from images. In this scenario the $\x$ vectors are the images and the $\y$ vectors represent the predicted locations of the person's joints. Therefore meaningful constraints are constraints on the relative predicted positions of the various joints. We will refer to them as \emph{Data Dependent} constraints because they are expressed in the output space of the network, and thus explicitly take into account the input vectors.   

Formally, let us consider a set of $N_C$ such constraints $\{C_j(\cdot)\}_{j=1}^{N_C}$. 
We seek to enforce all of these constraints on each input vector. However, accurately predicting the output vectors of the elements of $\dataset$ already implicitly satisfies the constraints, assuming that the training samples satisfy them. Therefore, following~\cite{Varol12a}, 
we introduce a set of unlabeled points $\cdataset = \{\x'_k\}_{k=1}^{|\cdataset|}$,  which may or may not share elements with~$\dataset$. The hard-constraint optimization problem of Eq.~\ref{eq:LossC} can then be formulated as solving 
\begin{equation}
    \min_{\w} R(\w) \quad \st \;  C_{jk}(\w) = 0 \quad \forall j \leq N_C, \forall k \leq |\cdataset| \; , \label{eq:constraintOpt1}
\end{equation}
where $C_{jk}(\w) = C_j(\phi(\x'_k; \w))$.

The corresponding soft-constraint problem tipically becomes solving 
\begin{equation}
    \min_{\w} R(\w) + \sum_j  \lambda_j \left(\sum_{k} C_{jk}(\w)^2 \right ) \;,    
    \label{eq:soft}
\end{equation}
where the $\lambda_j$ parameters are positive scalars that control the  relative contribution of each constraint. This is a compromise between minimizing the loss and satisfying the constraints. Too small a value of $\lambda_j$ will result in the corresponding constraint being ignored, and too large a one in the other terms, especially the loss, being ignored. 

The soft-constraint problem of Eq.~\ref{eq:soft} can be solved using techniques that have now become standard in the Deep Learning literature, such as SGD with momentum or using the Adam solver~\cite{kingman2014}. Even though constrained SGD methods have been proposed~\cite{Gasso11}, they have not been investigated in the Deep Learning context where the dimension of the vector $\w$ is very large, which is what we will do in the following section.


\section{Dealing with Millions of Variables}
\label{sec:KKT_million}

In this section, we first introduce standard Lagrangian methods to constrained optimization. Unfortunately, they cannot be directly used for our problem because

\begin{itemize}
 
 \item The parameter vector $\w$ of Eq.~\ref{eq:constraintOpt1}  has a large dimensionality, often in the order of millions or tens of millions.
 
 \item The constraints are not always guaranteed to be independent and might even be slightly incompatible, which introduces numerical instabilities. 
 
 \item The training datasets $\dataset$ and $\cdataset$ of Eq.~\ref{eq:constraintOpt1} are usually large as well. 
 
\end{itemize}

In the remainder of this section, we address each one of these issues in turn. In this work, we focus on equality constraints, but we will argue below that our approach could naturally extend to inequality ones. 

\subsection{Karush-Kuhn-Tucker Conditions}
\label{sec:KKT}

An effective way to solve a constrained optimization problem such as the one of Eq.~\ref{eq:constraintOpt1}  is to find a stationary point of the Lagrangian
\begin{gather}
    \min_{\w}\max_{\Lambda} \lagrangian(\w, \Lambda),   \label{eq:constraintHard} \\
   \mbox {with }  \lagrangian(\w, \Lambda) = R(\w) + \Lambda^T\C(\w), \nonumber
\end{gather}
where $\lagrangian$ is the Lagrangian and $\Lambda$ the $(|\cdataset| \cdot N_C)$-dimensional vector of {\it Lagrange multipliers}~\cite{Fletcher87}. They play the same role as the $\lambda_j$ of Eq.~\ref{eq:soft} but do not need to be set by hand. Since neither the loss nor the constraints are convex or linear, given an initial value $\w_0$  of $\w$, one computes an increment $\dw$ and the value of $\Lambda$ by solving a linearized version of Eq.~\ref{eq:constraintHard}, replacing $\w$ by  $\w+\dw$, and iterates. 

If the loss $L$ is a generic differentiable function, the increment $\dw$ is normally computed at each iteration by solving the linear system 
\begin{equation}
    \label{eq:linear_system1}
    \begin{bmatrix}
        \eta{}I & {\pdv{\C}{\w}}^T \\
        \pdv{\C}{\w} & 0 \\
    \end{bmatrix}
    \begin{bmatrix}
        \dw \\
        \Lambda
    \end{bmatrix} =
    \begin{bmatrix}
        -\pdv{R}{\w} \\
        -C(\w)
    \end{bmatrix} \; ,
\end{equation}
which amounts to projected gradient descent. The projection is performed onto the hyperplanes of linearized constraints. The linear system of Eq.~\ref{eq:linear_system1}  is derived from the Karush-Kuhn-Tucker (KKT) conditions that $\dw$ and $\Lambda$ must satisfy. 

When the risk~$R$ is a sum of squared residuals, which is typical in regression problems, it is generally more effective to replace the projected gradient by Gauss-Newton~(GN) or Levenberg-Marquardt~(LM). The iteration scheme remains the same but the KKT conditions become
\begin{equation}
    \label{eq:linear_system_lm}
    \begin{bmatrix}
        \mathbf{J}^T\mathbf{J} + \eta I & {\pdv{\C}{\w}}^T \\
        \pdv{\C}{\w} & 0 \\
    \end{bmatrix}
    \begin{bmatrix}
        \dw \\
        \Lambda
    \end{bmatrix} =
    \begin{bmatrix}
        -\mathbf{J}^T\mathbf{r}(\w_t) \\
        -C(\w_t)
    \end{bmatrix},
\end{equation}
where $\mathbf{r}$~is the vector of residuals and $\mathbf{J} = \pdv{\mathbf{r}}{\w}$
its Jacobian matrix evaluated at~$\w_t$. The full derivation can be found in~\cite{Fua10}.

\subsection{Krylov Subspace Methods} 

To perform the minimization, we must solve at each iteration a linear system of the form $\B\v = \b$ where the exact form of $\B$ and $\b$ depends on whether we enforce the KKT conditions of Eq.~\ref{eq:linear_system1} or~\ref{eq:linear_system_lm}. In either case,  the immense dimensionality of~$\w$ precludes explicitly computing or storing the matrix~$\B$. Instead, we solve the linear system using a Krylov subspace method~\cite{Krylov1931}.

Krylov subspace methods solve linear systems by
iteratively computing elements of the base of the Krylov subspace and finding
approximate solutions of the linear system in that subspace. The main advantage
of these methods is that they do not need an explicit representation of~$\B$.
Instead, the user has to provide a function that receives a vector~$\mathbf{v}$
and returns the matrix-vector product~$\B\mathbf{v}$. This
is usually much faster and less memory consuming than explicitly computing~$\B$.

Computing matrix-vector products with the matrices defined in 
Eqs.~\ref{eq:linear_system1} and~\ref{eq:linear_system_lm} involves
products of Jacobian matrices with vectors, which are essentially directional
derivatives. The \emph{Pearlmutter trick}~\cite{pearlmutter1994,vinyals2012}
is a well-known technique that
computes Jacobian-times-vector and vector-times-Jacobian products leveraging
the backpropagation procedures
common in neural networks. We call \emph{R-op} to the Jacobian-times-vector operation,
and we will write~$\Rop(\mathbf{f}, \w, \v)$ to express the backpropagation procedure that
computes the directional derivative of the multidimensional
function~$\mathbf{f}$ with respect to~$\w$ evaluated in the
direction~$\mathbf{v}$. That is,
\begin{equation}
    \Rop(\mathbf{f}, \w, \mathbf{v}) \equiv \pdv{\mathbf{f}}{\w}\mathbf{v}.
\end{equation}
Similarly, we call~\emph{L-op} the backpropagation procedure that computes
the \emph{adjoint directional derivative}, \ie, the vector-times-Jacobian product. That is,
\begin{equation}
    \Lop(\mathbf{f}, \w, \mathbf{v}) \equiv \mathbf{v}^T\pdv{\mathbf{f}}{\w}.
\end{equation}
\emph{R-op} and \emph{L-op} are available as GPU operators in many deep
learning packages such as~\emph{Theano}~\cite{bergstra2010}.

Given that~\emph{R-op} and \emph{L-op} are available, a Krylov subspace method
can solve the linear systems of Eq.~\ref{eq:linear_system1} and
Eq.~\ref{eq:linear_system_lm} using Algorithms~\ref{alg:compute_bv}
and~\ref{alg:compute_bv_lm}, respectively. Algorithm~\ref{alg:compute_bv} computes the
matrix-vector product for the matrix of Eq.~\ref{eq:linear_system1}, and
Algorithm~\ref{alg:compute_bv_lm} for the matrix of
Eq.~\ref{eq:linear_system_lm}.

\begin{algorithm}
\caption{Compute~$\B\v$ to solve Eq.~\ref{eq:linear_system1}}
\label{alg:compute_bv}
\begin{algorithmic}
    \Require Vector~$\v$
    \Ensure Product~$\B\v$
        \State $\v_1, \v_2 \gets$ Split $\v$ in two at position~$N_P$
        \State $\mathbf{a} \gets \eta\v_1 + \Lop(\C, \w, \v_2)$
        \State $\mathbf{b} \gets \Rop(\C, \w, \v_1)$
        \State \Return Concatenation of $\mathbf{a}$ and $\mathbf{b}$
\end{algorithmic}
\end{algorithm}

\begin{algorithm}
\caption{Compute~$\B\v$ to solve Eq.~\ref{eq:linear_system_lm}}
\label{alg:compute_bv_lm}
\begin{algorithmic}
    \Require Vector~$\v$
    \Ensure Product~$\B\v$
        \State $\v_1, \v_2 \gets$ Split $\v$ in two at position~$N_P$
        \State $\mathbf{Jv} \gets \Rop(\mathbf{r}, \w, \v_1)$
        \State $\mathbf{JJv} \gets \Lop(\mathbf{r}, \w, \mathbf{Jv})$
        \State $\mathbf{a} \gets \mathbf{JJv} + \eta\v_1 + \Lop(\C, \w, \v_2)$
        \State $\mathbf{b} \gets \Rop(\C, \w, \v_1)$
        \State \Return Concatenation of $\mathbf{a}$ and $\mathbf{b}$
\end{algorithmic}
\end{algorithm}

\subsection{Krylov Subspace Methods for Ill-conditioned Constraints}
\label{sec:krylov}

\begin{table*}
\begin{center}
\begin{tabular}{r|llll}
                        & Squared       & Symmetric     & Pos-definite     & Full rank \\
\hline
Conjugate gradient      & Yes           & Yes           & Yes               & Yes \\
Symmetric LQ~\cite{paige1975} & Yes     & Yes           & No                & Yes \\
\textbf{Minimal residual}~\cite{paige1975} & Yes & Yes           & No                & No \\
\textbf{Minimal residual + QLP}~\cite{choi2011}& Yes & Yes       & No                & No \\
Generalized minimal residual~\cite{saad1986} & Yes & No & No                & No \\
Biconjugate gradient    & No            & N/A           & N/A               & Yes \\
\end{tabular}
\end{center}
\caption{Krylov subspace methods for solving a linear system~$\B\v = \b$ and
their requirements for the matrix~$\B$.}
\label{tab:krylov}
\end{table*}

Although Krylov subspace methods allow us to solve high-dimensional
 linear systems, the solution might still suffer from numerical instabilities. In particular, the matrices of
Eqs.~\ref{eq:linear_system1} and~\ref{eq:linear_system_lm} are usually singular
or ill-conditioned as a consequence of incompatible or repeated constraints, which typically arise when linearizing constraints for a large number of samples.
As shown in Table~\ref{tab:krylov}, different Krylov subspace methods impose different conditions on the matrix~$\B$, and not all of them can handle ill-conditioned matrices.
Since our matrices are squared, symmetric, not positive-definite, 
and, crucially, ill-conditioned, we chose the \emph{minimal
residual method with QLP decomposition} (MINRES-QLP)~\cite{choi2011},  an improved version of the older (MINRES)~\cite{paige1975}, 
as our Krylov subspace method. 

\subsection{Stochastic Active Constraints}
\label{sec:active}

While Krylov subspace methods reduce the
memory required to solve Eq.~\ref{eq:linear_system1}, the size of the system
still depends on the number of constraints imposed, which is in the order of
the number of training samples. A simple method for
reducing this size consists of selecting a
subset~$\hat{\cdataset}\subset\cdataset$ of \emph{active constraints} from the unlabeled dataset
at each iteration, in a similar way
as stochastic gradient descent does with the labeled dataset. 

In some of our experiments, instead of randomly sampling~$\hat{\cdataset}$, we look for a subset
of elements with the largest violation of the constraint functions. That is,
\begin{eqnarray}
    \label{eq:mining}
    \hat{\cdataset} =& \displaystyle\argmax_{\hat{\cdataset}\subset\cdataset} &
        \sum_{\x'\in\hat{\cdataset}} \median_j \left|C_j(\phi(\x';\w_t))\right| \\
        & \st & |\hat{\cdataset}| < N_{\hat{\cdataset}}. \nonumber
\end{eqnarray}
This corresponds to \emph{hard sample mining} for constraints. It is easily extended to {\it inequality} constraints by treating them as equality constraints when they are violated and ignoring them otherwise. The active constraints are then chosen among the violated ones.

\subsection{Constrained Adam}
\label{sect:adam}

The step computed in Eq.~\ref{eq:linear_system1} is a projected gradient. Pure
gradient descent has known issues with~DNNs: it is sensitive to the chosen
learning rate, slow and prone to fall in bad local minima. Methods like
momentum gradient descent solve these problems by averaging the gradient over
several iterations, and others like Adagrad~\cite{duchi2010} or
Adadelta~\cite{zeiler2012} compute a parameter-wise learning rate based on the
previous updates of every parameter. Adam~\cite{kingman2014} combines momentum
with a parameter-wise learning rate.

We extend our constrained training method with an Adam-like step. At every iteration, we
update first and second order moment vectors~$\mathbf{m}$ and~$\mathbf{v}$ as
\begin{eqnarray}
    \mathbf{m}_{t+1} &=& \beta_1 \mathbf{m}_{t} + (1 - \beta_1) \pdv{R}{\w}, \\
    \mathbf{v}_{t+1} &=& \beta_2 \mathbf{v}_{t} + (1 - \beta_2) \left(\pdv{R}{\w}\right)^{\circ 2},
\end{eqnarray}
for given hyperparameters~$\beta_1$ and~$\beta_2$ and partial derivatives evaluated
at~$\w_{t}$. $\beta_1$ and~$\beta_2$ control the exponential decay rates of the vectors~$\mathbf{m}$ and~$\mathbf{v}$.
We set them to~$0.9$ and~$0.999$ respectively, as proposed in~\cite{kingman2014}.
The linear system of the constrained version of Adam uses the
vector~$\mathbf{m}$ as the current gradient
and~$\sqrt{\mathbf{v}}$ as the parameter-wise learning rate. This yields
\begin{equation}
    \label{eq:linear_system_adam}
    \begin{bmatrix}
        \eta f \diag\left(\sqrt{\mathbf{v}_{t+1}} + \epsilon\right) & {\pdv{\C}{\w}}^T \\
        \pdv{\C}{\w} & 0 \\
    \end{bmatrix}
    \begin{bmatrix}
        \dw \\
        \Lambda
    \end{bmatrix} =
    \begin{bmatrix}
        -\mathbf{m}_{t+1} \\
        -C(\w_t)
    \end{bmatrix},
\end{equation}
with the correction factor~$f=\dfrac{\sqrt{1 - \beta_2^{t+1}}}{1-\beta_1^{t+1}}$.
Algorithm~\ref{alg:compute_bv} can be easily modified to compute matrix-vector
products with this constrained Adam matrix.


\section{Experiments}
\label{sec:experiments}

\begin{figure}
    \[
    \begin{array}{cccc}
        \includegraphics[width=0.23\linewidth]{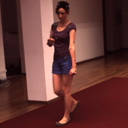} &
        \includegraphics[width=0.23\linewidth]{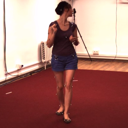} &
        \includegraphics[width=0.23\linewidth]{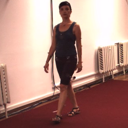} &
        \includegraphics[width=0.23\linewidth]{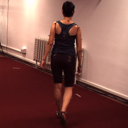} \\
        \includegraphics[width=0.15\linewidth,trim={8.5cm 4.5cm 7.5cm 5cm},clip]{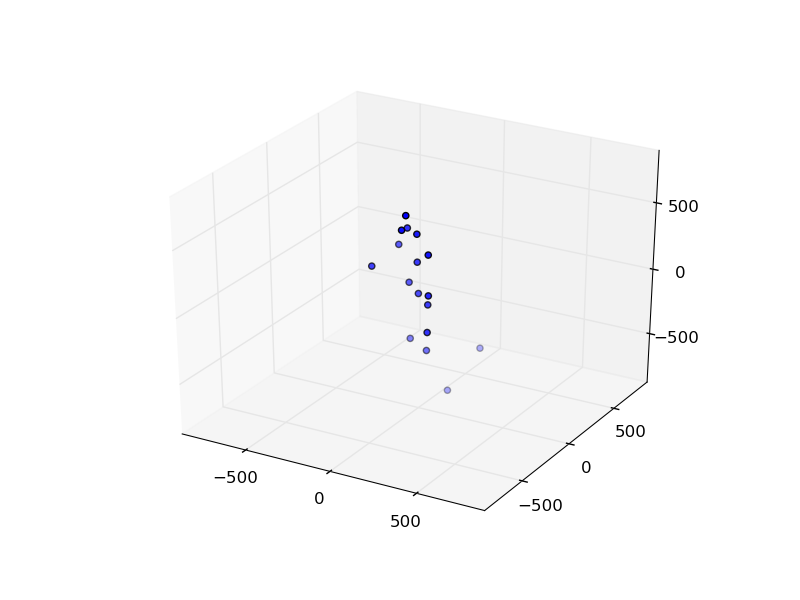} &
        \includegraphics[width=0.15\linewidth,trim={8.5cm 4.5cm 7.5cm 5cm},clip]{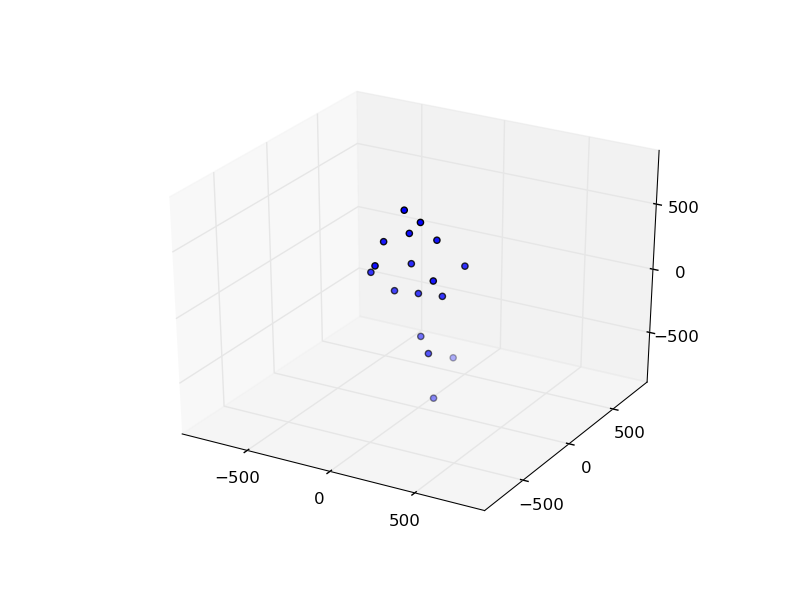} &
        \includegraphics[width=0.15\linewidth,trim={8.5cm 4.5cm 7.5cm 5cm},clip]{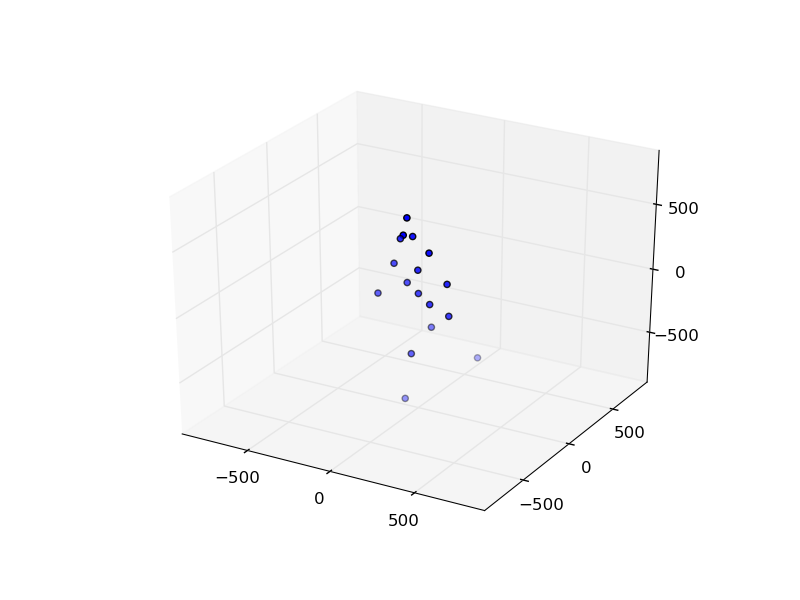} &
        \includegraphics[width=0.15\linewidth,trim={8.5cm 4.5cm 7.5cm 5cm},clip]{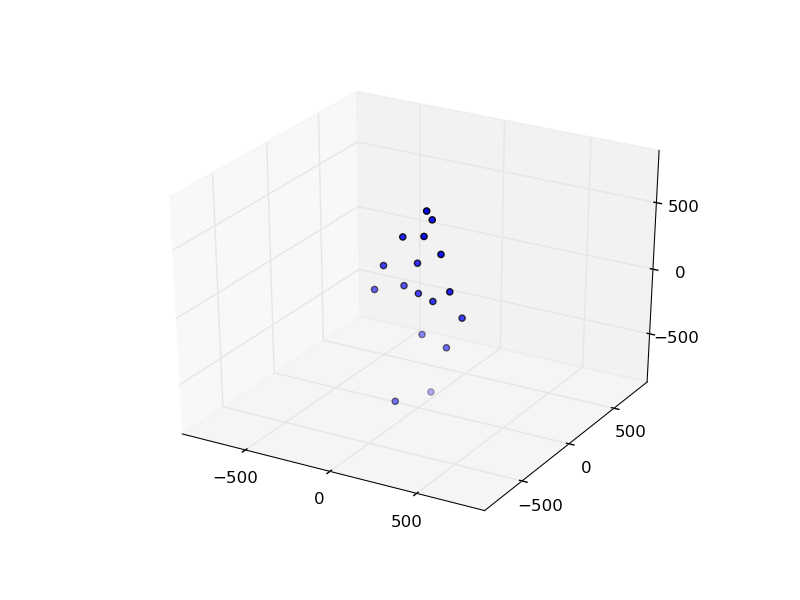}
    \end{array}
    \]
    \vspace{-0.5cm}
    \caption{Samples from the Human3.6m dataset. Our models are trained  to map from $128\times 128$~color images, as those of the top row, to the 3D~pose  of 17~joints, as shown in the bottom row.}
    \label{fig:training_data}
\end{figure}

To compare the respective merits of soft and hard constraints when used in conjunction with Deep Networks, we used the well-known and important problem of 3D human pose estimation from a single view. We performed our experiments on the \emph{Walking} sequence of the \emph{Human3.6m} dataset~\cite{Ionescu2011,Ionescu2014}. It consists of 162008~color images that were cropped to~$128\times 128$~pixels. The ground-truth data for each image corresponds to the 3D coordinates of 17~joints of the human body as shown if Figure~\ref{fig:training_data}. We split the dataset into 80\%-20\% subsets for training and testing, respectively.

To define the network $\phi$, we used an  architecture that is a composition of convolutional
layers, max-pooling, ReLU non-linearities and fully-connected layers in the following order: 
\begin{eqnarray}
    \phi &=& \mathcal{A}_{51}\circ\mathcal{R}\circ\mathcal{A}_{1024}\circ\mathcal{R}\circ\mathcal{A}_{1024}\circ\mathcal{C}_{8,3\times3} \\
    && \hbox{}\circ\mathcal{R}\circ\mathcal{P}_{2\times2}\circ\mathcal{C}_{8,5\times5}\circ\mathcal{R}\circ\mathcal{P}_{4\times4}\circ\mathcal{C}_{8,5\times5}, \nonumber
\end{eqnarray}
where $\mathcal{C}_{a,b}$~is a convolutional layer of $a$~filters with size~$b$, $\mathcal{P}_{a}$~is a max-pooling layer with block size~$a$, $\mathcal{R}$~is the ReLU non-linearity and $\mathcal{A}_a$~is an affine transformation, or fully connected layer, with $a$~outputs. Layers of type~$\mathcal{C}$ and~$\mathcal{A}$ are parameteric, but we omitted the parameters for notational simplicity. Even though this architecture is not state-of-the-art anymore, it still gives reasonably good results while remaining simple enough to perform numerous experiments. Given that are we not interested in absolute performance but in comparing performances obtained using different approaches to imposing constraints, this is what we need. 

We took the supervised  loss~$L$ of Eq.~\ref{eq:LossU}
to be the squared distance between the ground-truth and the prediction for each joint. 
This can be expressed as
\begin{equation}
    L(\hat{\y}_i, \y_i) = \dfrac{1}{17\cdot3}\|\hat{\y}_i - \y_i\|^2 \; ,
    \label{eq:distLoss}
\end{equation}
where each $\y_i$ represents the true 3D position of one the 17 joints and $\hat{\y}_i$ the position predicted by the network from the corresponding image. 

We first trained the network~$\phi$ by minimizing this loss without imposing either hard or soft constraints. We ran 100 epochs of the Adam solver~\cite{kingman2014}
with a minibatch size of 128. We take the model that reached the smallest error on the validation set as our \emph{unconstrained} model.



We then defined six symmetry constraints: Equal lengths of left and right arms, forearms, legs and calves, equal distance from the two shoulders to the spine, and equal distance of the two hips to the spine. They are different from those of~\cite{Zhou17a}  but similar in spirit. The constraint functions are written as
\begin{eqnarray}
    \label{eq:symmetry_constraints}
    C_{jk} \equiv C_j(\hat{\y}_k) &=& \|\hat{\y}_{k,\joint(j, 1)} - \hat{\y}_{k,\joint(j, 2)}\| \\
    && \hbox{} - \|\hat{\y}_{k,\joint(j, 3)} - \hat{\y}_{k,\joint(j,4)}\| \; , \nonumber
\end{eqnarray}
where $j$ and~$k$ are the constraint and the sample indices, respectively.
$\hat{\y}_{k, \joint(j, m)}$~is
the predicted 3D position of the joint~$\joint(j, m)$ for data sample~$k$, and
$\joint(\cdot, \cdot)$~is an auxiliary function indexing the joints involved
in every constraint. Its complete definition is given in
Table~\ref{tab:constraint_joints}.
A constraint~$C_{jk}$ is therefore met when the distance between
joints~$\joint(j,1)$ and~$\joint(j,2)$ is equal to the distance between
$\joint(j,3)$ and~$\joint(j,4)$.

\begin{table*}
\begin{center}
\begin{tabular}{l|cccc}
        & $m=1$         & $m=2$         & $m=3$             & $m=4$ \\
\hline
$j=1$   & left shoulder & left elbow    & right shoulder    & right elbow \\
$j=2$   & left elbow    & left hand     & right elbow       & right hand \\
$j=3$   & left hip      & left knee     & right hip         & right knee \\
$j=4$   & left knee     & left heel     & right knee        & right heel \\
$j=5$   & chest         & left shoulder & chest             & right shoulder \\
$j=6$   & pelvis        & left hip      & pelvis            & right hip \\
\end{tabular}
\end{center}
\caption{Joints for the symmetry constraint functions of Eq.~\ref{eq:symmetry_constraints}.
Every cell contains the value~$\joint(j, m)$.}
\label{tab:constraint_joints}
\end{table*}

We trained the following constrained models in several different ways:
\begin{description}

    \item[Soft-SGD:] Standard stochastic gradient descent to minimize the loss of Eq.~\ref{eq:soft}. We tried different values of the $\lambda_j$ parameters of Eq.~\ref{eq:soft}, which we take to all be equal to each other, and of the learning rate~$\eta$.
    
    \item[Soft-Adam:] Same as \textbf{Soft-SGD}, but using the Adam solver instead. In this case the learning rate was fixed to~$\eta=10^{-3}$ because Adam is very insensitive to that choice.
        
    \item[Hard-SGD:] Our method using Eq.~\ref{eq:linear_system1} solved using MINRES-QLP. We tried different parameters for the learning rate $\eta$.
    
    
    \item[Hard-Adam:] Same as  \textbf{Hard-SGD}, but using our Adam version of Eq.~\ref{eq:linear_system_adam}.
    
\end{description}

\begin{figure*}
    \[
    \begin{array}{cc}
        \includegraphics[width=0.4\linewidth]{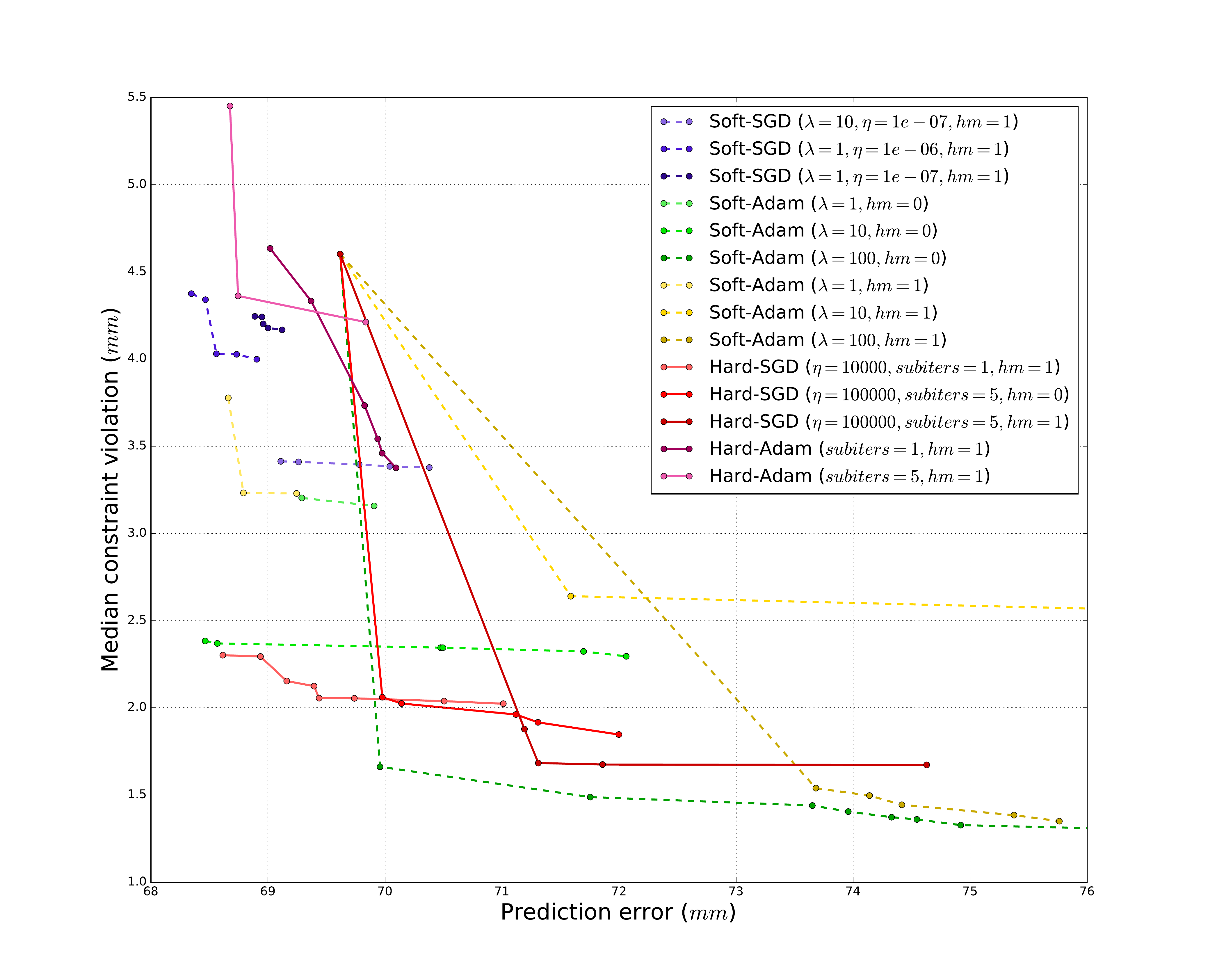} &
        \includegraphics[width=0.4\linewidth]{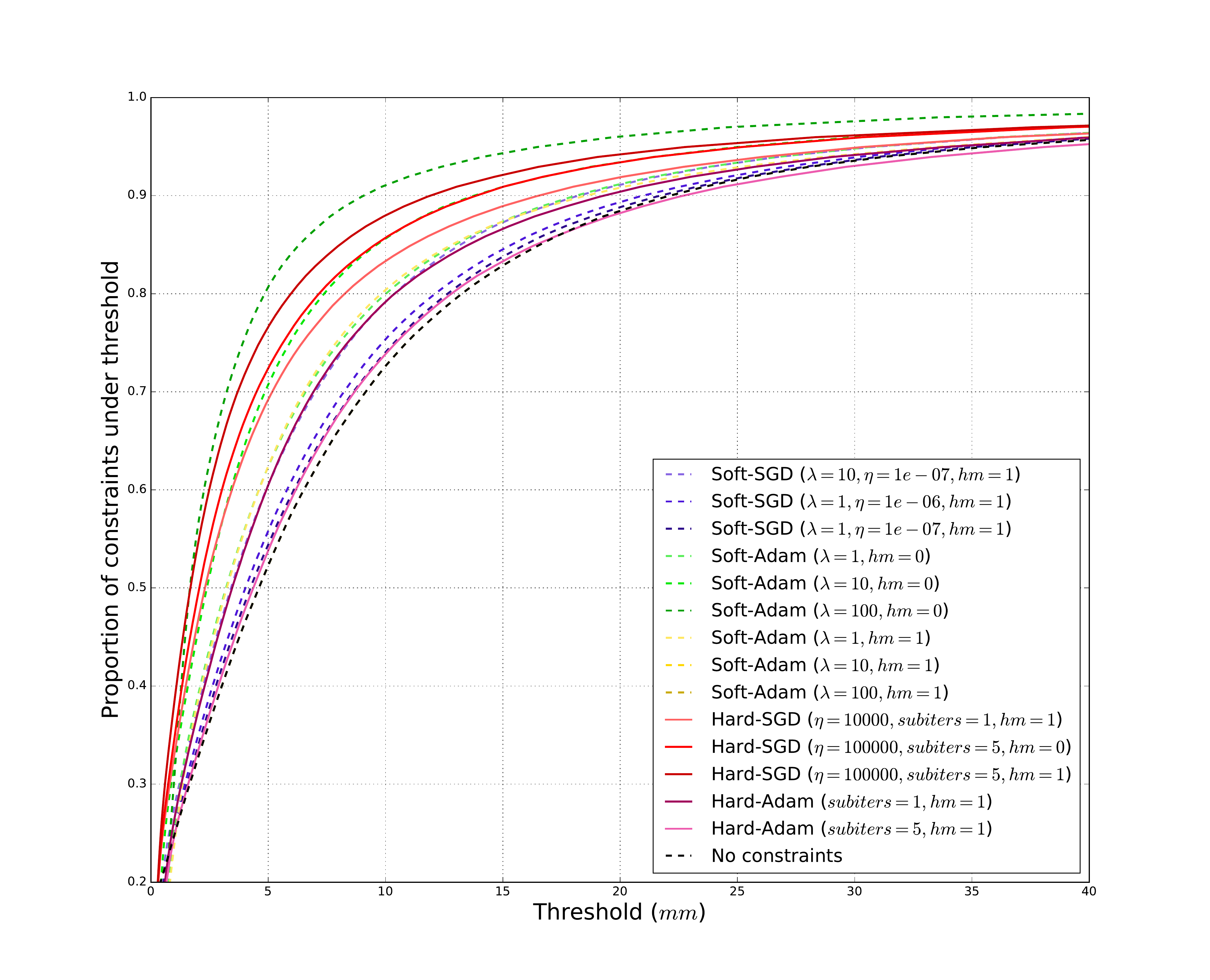} \\
        \textrm{(a)} & \textrm{(b)}\\
    \end{array}
    \]
    \vspace{-0.5cm}
    \caption{Comparison of soft and hard constraint methods for human pose estimation.
    (a)~Comparison of average errors vs.~median of absolute values of the constraints.
    Every dot represents a model. The large blue dot represents the unconstrained model. Lines connect models obtained with the same
    configuration but different number of training iterations.
    (b)~Proportion of constraints met vs.~threshold for different methods.
    In both plots the line style indicates
    the class of the method: solid lines for hard constraints and dashed lines for soft
    constraints.
    }
    \label{fig:hpe_results}
\end{figure*}

All constrained models were trained on the same training dataset $\dataset$ we used to learn the unconstrained model, using again minibatches of 128 samples.  Similarly, we used the same dataset $\cdataset$ on which we enforced the constraints for all the constrained models.
There, we used either random minibatches of 128~samples from~$\cdataset$ or 
performed hard constraint mining according to Eq.~\ref{eq:mining},
keeping only 16~samples at a time.
We used two metrics for evaluation purposes:
\begin{itemize}
\item {\bf Prediction error}. Mean of the Euclidean distances between
the predictions of the joint positions and the real positions, averaged over the
whole validation dataset:
\begin{equation*}
    \textrm{error}(\mathcal{Y}, \hat{\mathcal{Y}}) =
        \dfrac{1}{N} \sum_{i=1}^N \dfrac{1}{17} \sum_{m=1}^{17} \|\y_{i,m} - \hat{\y}_{i,m}\|.
\end{equation*}
\item  {\bf Median constraint violation}. Median of the absolute values of all~$C_{jk}$.
\end{itemize}
We report our comparative results in Fig.~\ref{fig:hpe_results}.  \textbf{Soft-Adam} yields the best results, closely followed by \textbf{Hard-SGD}. Generally speaking, the Hard-constraint methods perform similarly to the soft-constraint methods in both metrics, instead of being better as we had hoped. In particular, the hard-constraint methods do not enforce the constraints perfectly, not even on the training data. This is because we can only use a subset of the constraints at each iteration, as explained in Section~\ref{sec:active}. As a result, constraints that were satisfied at one iteration can stop being satisfied at the next.

The parameter-wise adaptive learning rate of Adam leads to convergence even when using high values of~$\lambda$ without causing numerical instabilities, which makes it relatively easy to 
choose $\lambda$. Conversely,  \textbf{Soft-SGD} is very sensitive to the choice of both the learning rate and $\lambda$ because it lacks the adaptability of Adam. Surprisingly, Adam does not perform well with hard-constraint methods. We speculate that this is a consequence of the Adam momentum not dealing well with the projection onto the linearized constraint surface at each iteration.


We implemented all the methods in Python with Theano~\cite{bergstra2010}
using the MINRES-QLP implementation from~\cite{Pascanu2013}, and ran the experiments
on an nVidia Titan~Z.
\textbf{Soft-SGD} and \textbf{Soft-Adam} required between $0.8$ and $1.6$~seconds per iteration (s/it)
respectively, while \textbf{Hard-SGD} required between~$3$\,s/it and~$32$\,s/it, depending on the properties
of the linear system solved at each update. \textbf{Hard-Adam} ran at $103$\,s/it.
As expected, hard-constraint techniques are on average between 10 and 100~times
slower than their soft-constraint counterparts. 


\section{Discussion}
\label{sect:discussion}

The human pose estimation experiments described in the previous section indicate that learning with hard constraints is computationally tractable, if more expensive than with soft constraints, and yields exploitable results. However, it does not guarantee either perfect satisfaction of the constraints, not even in the training data, or better performance than using soft constraints in conjunction with the Adam solver~\cite{kingman2014}, which is not what we expected to happen when we started working on this problem.

\subsection{Interpretation}

This could of course be a fluke of the specific network architecture and constraint set we used in our experiments. In this section, we argue here that  this not the case and points to a fundamental problem with imposing hard constraints on Deep Nets that needs to be addressed: Lagrangian methods applied to non-linear loss and constraint functions involve repeatedly linearizing and solving the KKT conditions of Eqs.~\ref{eq:linear_system1} or~\ref{eq:linear_system_lm}, which means solving linear systems to project the potential solution vectors onto the intersection of a set of hyperplanes. The Krylov subspace approach we introduced in Section~\ref{sec:krylov} has made those computationally tractable despite the enormous size of the linear systems while the MINRES-QLP approach~\cite{choi2011} to solving them delivers numerical stability even though the matrices are ill-conditioned. However, we are still left with the problem that we cannot enforce all the constraints all the time and had to resort to using a randomly chosen subset of constraints at each iteration. As already observed, this means that constraints that are satisfied at one iteration can stop being satisfied at the next. The problem is compounded by the fact that the constraints $C_j(f(\x'_k; \w))$ of Eq.~\ref{eq:constraintOpt1}  are {\it data-dependent}, meaning they depend not only on the parameter vector $\w$ but also of the data vector $\x'$ at which they are evaluated. This means that we cannot guarantee that their linearizations are linearly independent from each other, which makes the KKT systems even more ill-conditioned, making the work of MINRES-QLP even more difficult. We believe this to be the root-cause of our difficulties.

\subsection{Demonstration on a Synthetic Example}

To demonstrate this, we now introduce a simple synthetic problem with mildly incompatible constraints, but with as many variables as in the previous Deep Nets. We then show that it gives rise to the same behavior as the one we reported in Section~\ref{sec:experiments}, that is, treating the constraints as hard constraints can be made to work but, in the end, treating them as soft constraints remains preferable if we have to work with different constraint subsets at each iteration.  

Let us consider the hard-constraint problem of solving
\begin{equation}
    \label{eq:synthetic_hard}
    \min_{\w} \dfrac{1}{2}\|\w - \x_0\|^2\;\st \|\w - \c_i\| - 10 = 0,\; 1\le i \le 200
\end{equation}
and the corresponding soft-constraint one of solving
\begin{equation}
    \label{eq:synthetic_soft}
    \min_{\w} \dfrac{1}{2}\|\w - \x_0\|^2 + \lambda \sum_{1\le i \le 200} \left(\|\w - \c_i\| - 10\right)^2 \; ,
\end{equation}
when $\w$, $\x_0$, and the $\c_i$ are vectors of dimension~$d$. In other words, we look for a vector~$\w$ that is closest to a fixed vector~$\x_0$, while being at the intersection of $200$~different hyperspheres. Each one is centered at $\c_i$, which we take to be normally distributed around the origin with a variance of~$0.01$ that is two order of magnitude smaller than the hypersphere radius. In other words, all these constraints are very similar but slightly incompatible. For the soft-constraint problem, we take $\lambda$ to be~$100$.

\begin{figure*}
    \[
    \begin{array}{ccc}
        \includegraphics[width=0.31\linewidth]{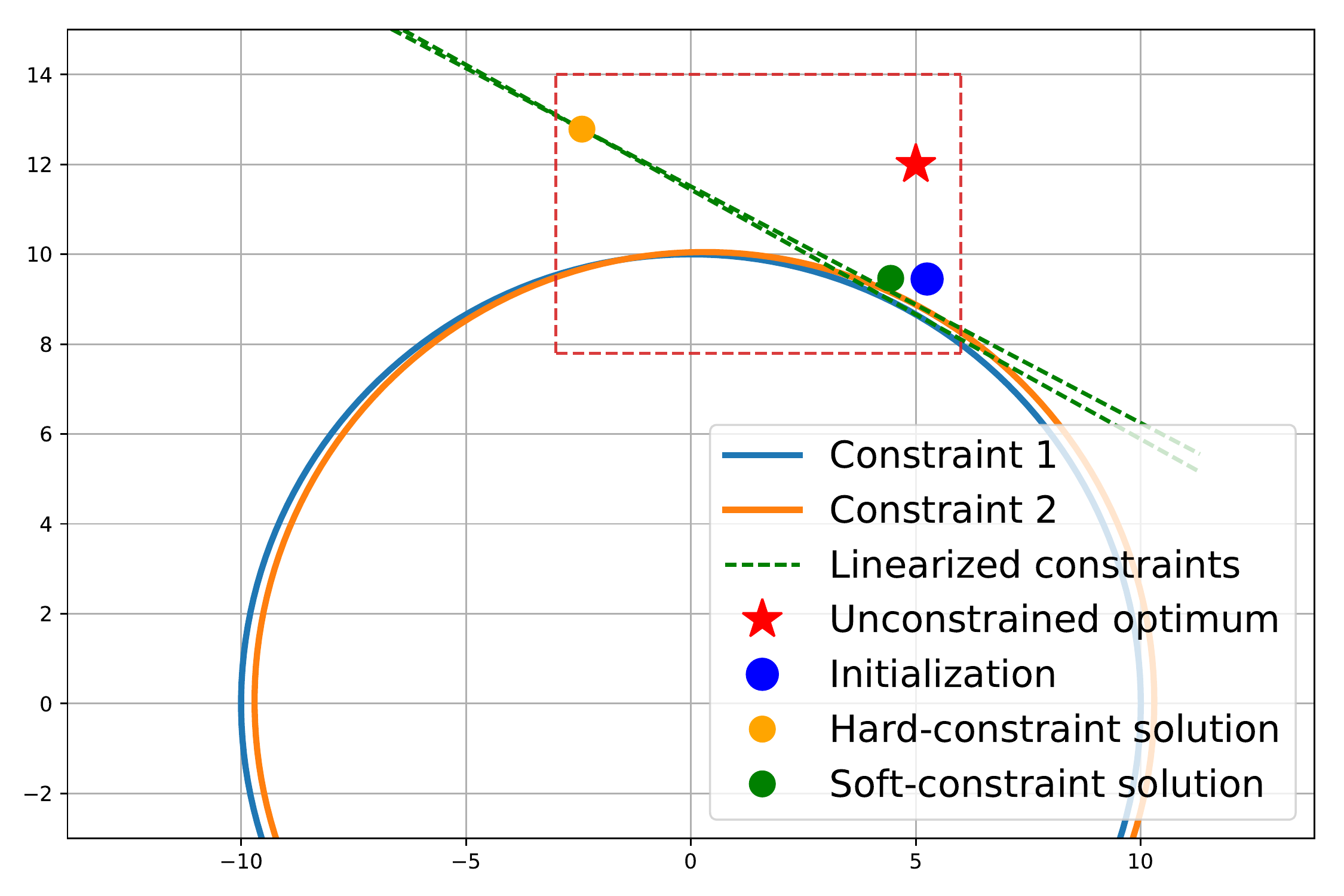} &
        \includegraphics[width=0.31\linewidth]{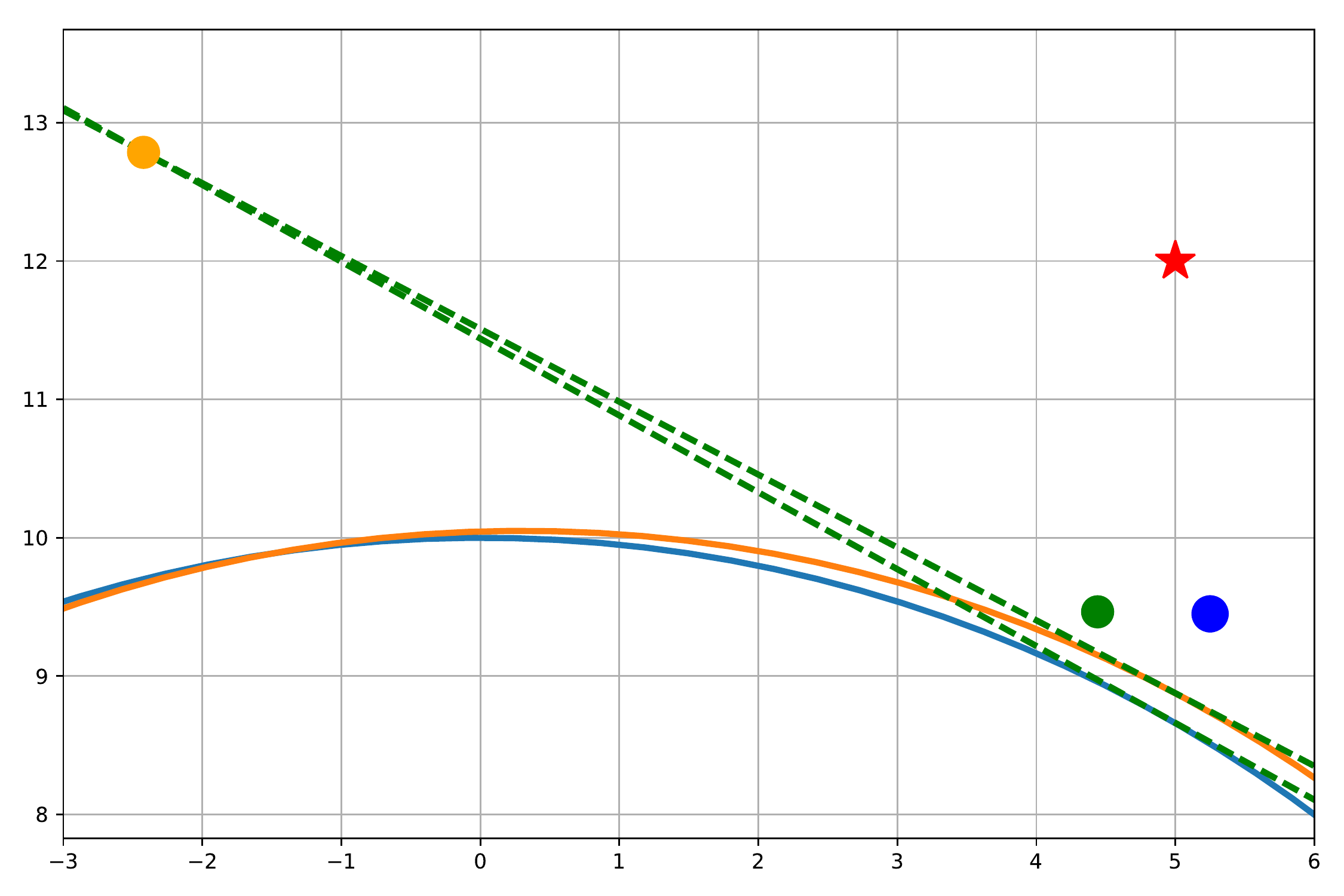} &
        \includegraphics[width=0.31\linewidth]{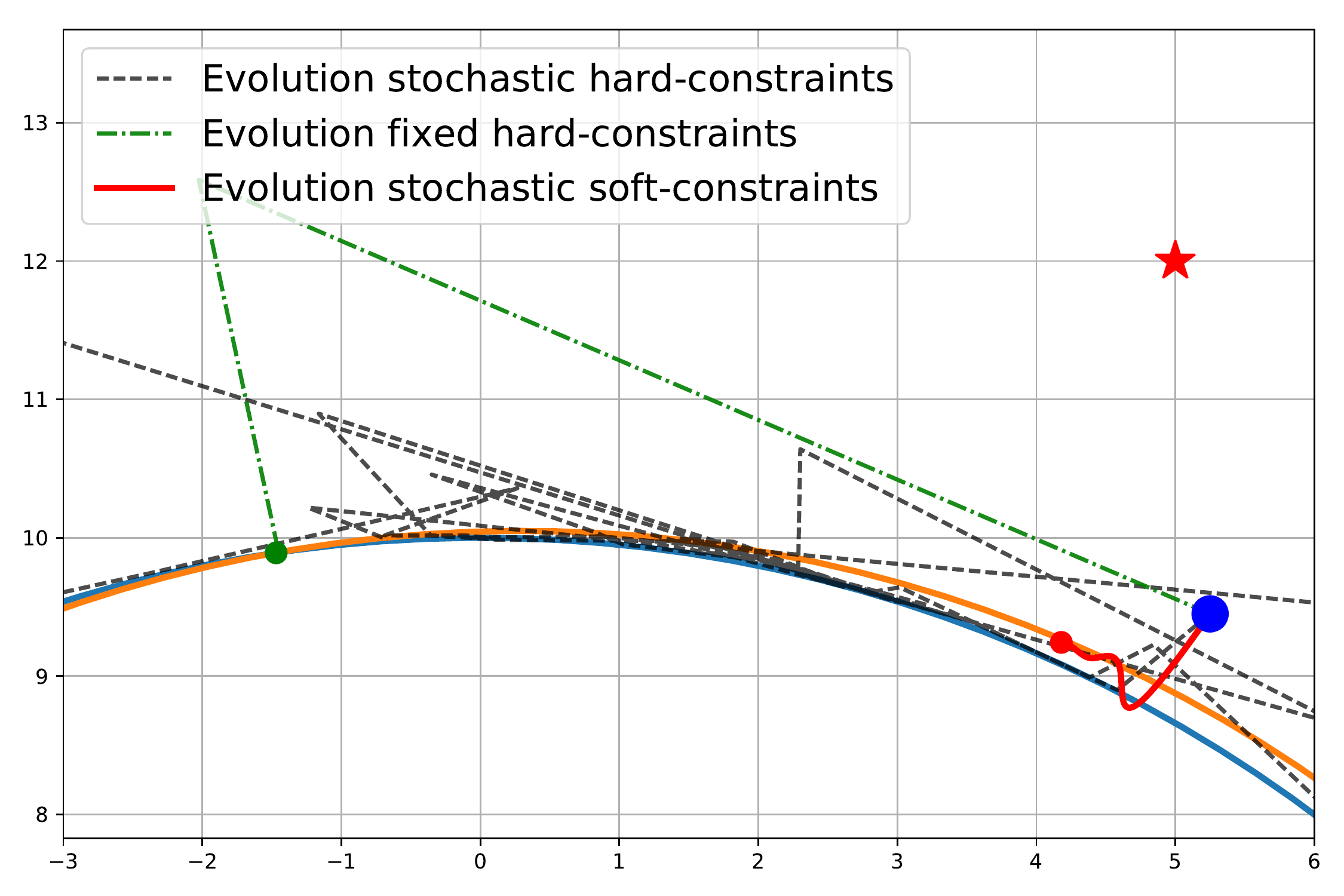} \\
        \textrm{(a)} & \textrm{(b)} & \textrm{(c)}
    \end{array}
    \]
    \vspace{-0.5cm}
    \caption{Synthetic experiment with only 2~active constraints. (a) One iteration starting from the blue dot and leading to either the green dot when using soft constraints or the yellow one when using hard constraints. (b) Zoomed in version of (a). (c) The optimization path for many iterations starting from the blue dot, when using soft constraints (red line), the same two hard constraints all the time (green dotted line), and different pairs of constraints at every iteration (back dotted line). The latter is totally erratic. }
    \label{fig:spheres2d_example}
\end{figure*}

Fig.~\ref{fig:spheres2d_example}(a,b) depicts a single iteration where only two constraints are active and $d=2$. We choose these two constraints to be very similar but slightly incompatible.  Linearizing them therefore results in hyperplanes---lines in this case---that are almost parallel and intersect far away. In a sense, this a worst-case scenario for hard-constraints, with the linear manifolds intersecting far from the original constraints.  As a result, imposing the KKT conditions by solving the linear system of Eq.~\ref{eq:linear_system1}  sends the candidate solution towards the intersection of the two hyperplanes and far away from the real constraint surfaces. If we repeated this operation using the {\it same} two constraints, $\w$ would eventually converge to the one feasible point at the intersection of the two constraint circles nevertheless. However, as discussed in Section~\ref{sec:active}, the set of active constraints changes at each iteration, resulting in an erratic behavior of $\w$ as the optimization progresses, as shown in Fig.~\ref{fig:spheres2d_example}(c).

\begin{figure*}
    \[
    \begin{array}{cc}
        \includegraphics[width=0.45\linewidth]{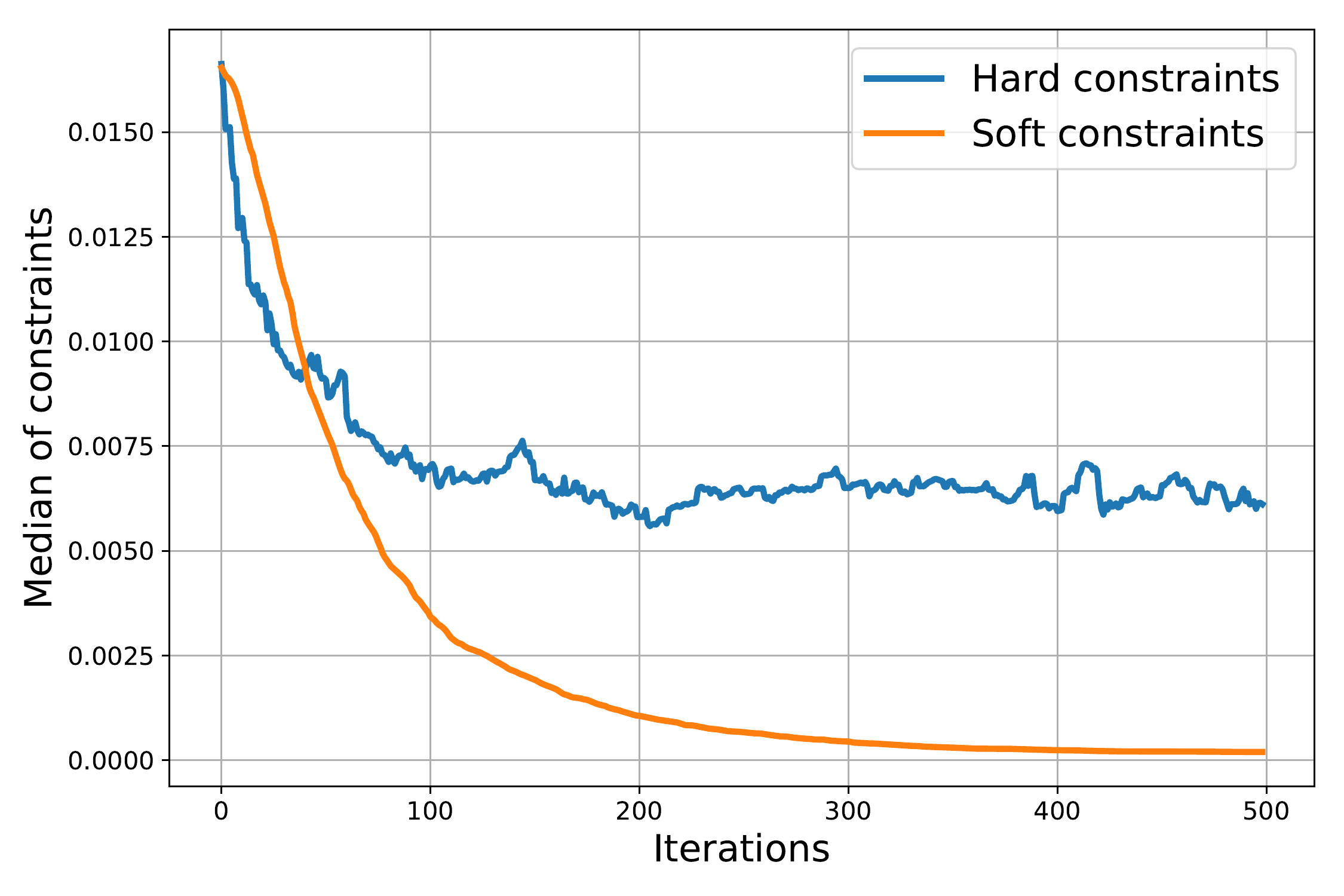} &
        \includegraphics[width=0.45\linewidth]{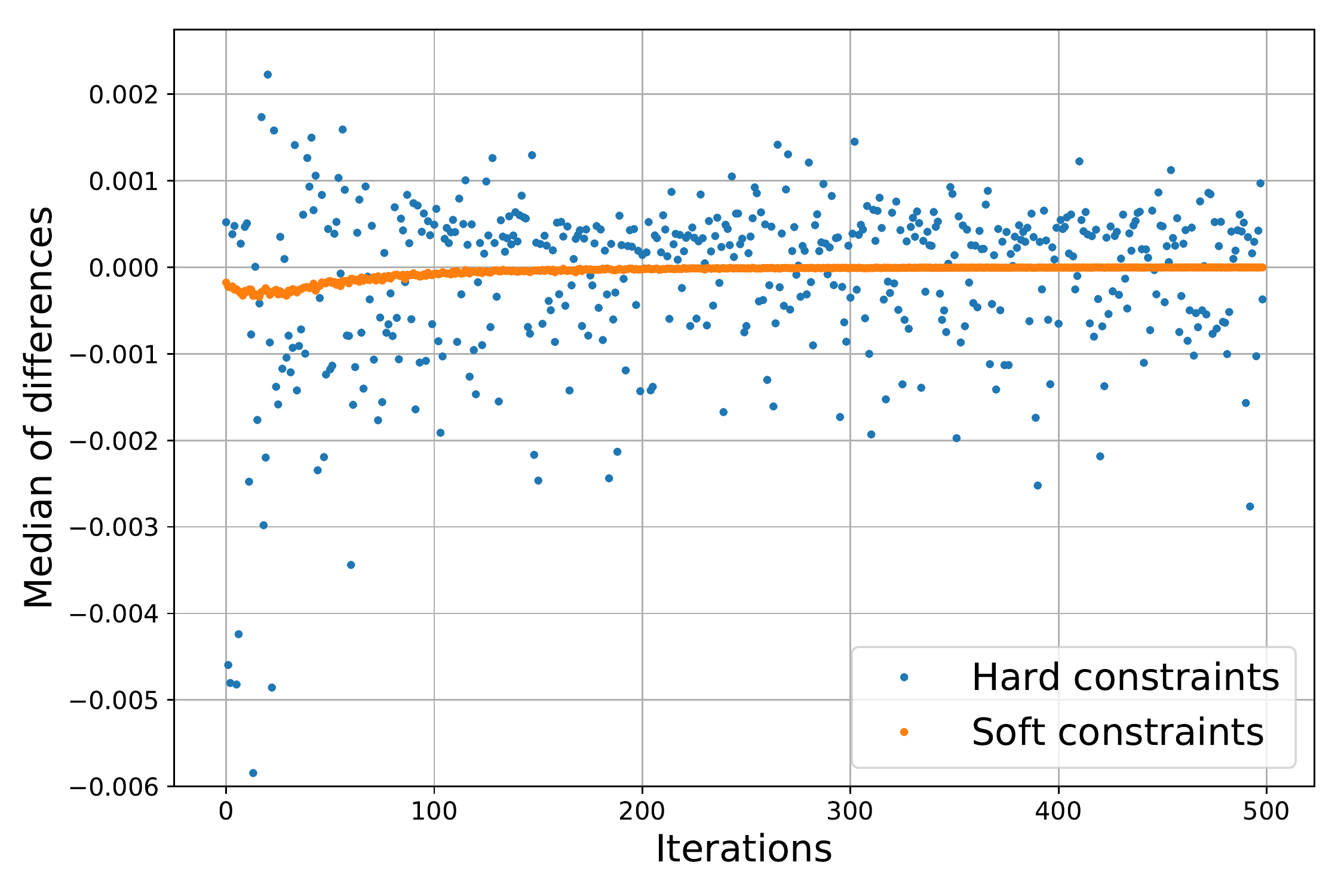} \\
        \textrm{(a)} & \textrm{(b)}
    \end{array}
    \]
    \vspace{-0.4cm}
    \caption{Hypersphere experiment of Eq.~\ref{eq:synthetic_hard}
    and~\ref{eq:synthetic_soft}. (a)~Median of the absolute value of all the
    200~constraint functions at every iteration. (b)~Differences between the absolute
    value of the \emph{active} constraints before and after the update step at
    every iteration. A positive value indicates that the update step deteriorated
    the quality of the active constraints. Note that these are values of the original constraints,
    as opposed to the linearized ones.}
    \label{fig:spheres_comparison}
\end{figure*}

To simulate a problem of a size roughly comparable to that of a Deep Learning one,  we solved the hard- and soft-constraint problems of Eq.~\ref{eq:synthetic_soft} with $d=1e6$. To enforce the hard constraints, we used the method of Eq.~\ref{eq:linear_system1}. For the soft ones, we used stochastic gradient descent without momentum. In both cases we used a subset of 20~randomly chosen active constraints at every iteration. Fig.~\ref{fig:spheres_comparison}(a) depicts the resulting evolution of the median of the absolute value of the 200~constraints over 500~iterations. In this simple experiment, not only do soft constraints perform much better than hard constraints, but their behavior is smoother and more stable.

To quantify how common it is for KKT updates to make the constraints {\it less} well satisfied as opposed to better satisfied, which is what they are designed to do, we computed the value of the active constraints before and after updating the parameters. In Fig.~\ref{fig:spheres_comparison}(b) we plot the differences of the median of the absolute value of the constraints before and after updating. Negative values therefore indicate that the active constraints are better satisfied after update than before. Conversely positive values denote a degradation. In the soft-constrain case, there is initially a steady improvement until the algorithm stabilizes. By contrast, in the hard-constraint case, the behavior is far more unpredictable and chaotic, while eventually yielding a similar value of the loss.

\comment{}


\section{Conclusion}

We have shown that it is practical train a Deep Network architecture while imposing hard constraints on its output. To this end, we have developed a Lagrangian method that relies on a Krylov subspace approach to solving the resulting very large linear systems. Unfortunately, our experiments have shown that our approach performs well, but not better than using a soft-constraint approach as we had hoped. 

We attribute this negative result the fact that, to keep the computational cost within reasonable limits, we had to choose a new subset of active constraints at every iteration instead of using all constraints all the time, thus forgetting the effect of all previously used ones. This is clearly suboptimal as successful hard-constraint methods do not normally remove elements from the pool of active constraint, at least when dealing when equality constraints instead of inequality ones.  Implementing this in the Deep Network context is far from trivial, but we hope that this paper might inspire other researchers to look into it.

\clearpage
{\small
\bibliographystyle{ieee}
\bibliography{top,string,vision,optim,biomed,learning}
}

\end{document}